\PassOptionsToPackage{unicode}{hyperref}
\PassOptionsToPackage{hyphens}{url}
\PassOptionsToPackage{dvipsnames,svgnames,x11names}{xcolor}
\documentclass[
  12pt,
  letterpaper,
  DIV=11,
  numbers=noendperiod]{scrartcl}

\usepackage{amsmath,amssymb}
\usepackage{iftex}
\ifPDFTeX
  \usepackage[T1]{fontenc}
  \usepackage[utf8]{inputenc}
  \usepackage{textcomp} 
\else 
  \usepackage{unicode-math}
  \defaultfontfeatures{Scale=MatchLowercase}
  \defaultfontfeatures[\rmfamily]{Ligatures=TeX,Scale=1}
\fi
\usepackage{lmodern}
\ifPDFTeX\else  
\fi
\IfFileExists{upquote.sty}{\usepackage{upquote}}{}
\IfFileExists{microtype.sty}{
  \usepackage[]{microtype}
  \UseMicrotypeSet[protrusion]{basicmath} 
}{}
\makeatletter
\@ifundefined{KOMAClassName}{
  \IfFileExists{parskip.sty}{%
    \usepackage{parskip}
  }{
    \setlength{\parindent}{0pt}
    \setlength{\parskip}{6pt plus 2pt minus 1pt}}
}{
  \KOMAoptions{parskip=half}}
\makeatother
\usepackage{xcolor}
\makeatletter
\ifx\paragraph\undefined\else
  \let\oldparagraph\paragraph
  \renewcommand{\paragraph}{
    \@ifstar
      \xxxParagraphStar
      \xxxParagraphNoStar
  }
  \newcommand{\xxxParagraphStar}[1]{\oldparagraph*{#1}\mbox{}}
  \newcommand{\xxxParagraphNoStar}[1]{\oldparagraph{#1}\mbox{}}
\fi
\ifx\subparagraph\undefined\else
  \let\oldsubparagraph\subparagraph
  \renewcommand{\subparagraph}{
    \@ifstar
      \xxxSubParagraphStar
      \xxxSubParagraphNoStar
  }
  \newcommand{\xxxSubParagraphStar}[1]{\oldsubparagraph*{#1}\mbox{}}
  \newcommand{\xxxSubParagraphNoStar}[1]{\oldsubparagraph{#1}\mbox{}}
\fi
\makeatother

\usepackage{color}
\usepackage{fancyvrb}

\DefineVerbatimEnvironment{Highlighting}{Verbatim}{commandchars=\\\{\}}
\usepackage{framed}
\definecolor{shadecolor}{RGB}{241,243,245}
\newenvironment{Shaded}{\begin{snugshade}}{\end{snugshade}}

\newcommand{\BuiltInTok}[1]{\textcolor[rgb]{0.00,0.23,0.31}{#1}}
\newcommand{\CharTok}[1]{\textcolor[rgb]{0.13,0.47,0.30}{#1}}
\newcommand{\CommentTok}[1]{\textcolor[rgb]{0.37,0.37,0.37}{#1}}

\newcommand{\ControlFlowTok}[1]{\textcolor[rgb]{0.00,0.23,0.31}{\textbf{#1}}}

\newcommand{\DecValTok}[1]{\textcolor[rgb]{0.68,0.00,0.00}{#1}}

\newcommand{\FloatTok}[1]{\textcolor[rgb]{0.68,0.00,0.00}{#1}}

\newcommand{\ImportTok}[1]{\textcolor[rgb]{0.00,0.46,0.62}{#1}}

\newcommand{\KeywordTok}[1]{\textcolor[rgb]{0.00,0.23,0.31}{\textbf{#1}}}
\newcommand{\NormalTok}[1]{\textcolor[rgb]{0.00,0.23,0.31}{#1}}
\newcommand{\OperatorTok}[1]{\textcolor[rgb]{0.37,0.37,0.37}{#1}}

\newcommand{\PreprocessorTok}[1]{\textcolor[rgb]{0.68,0.00,0.00}{#1}}

\newcommand{\SpecialCharTok}[1]{\textcolor[rgb]{0.37,0.37,0.37}{#1}}
\newcommand{\SpecialStringTok}[1]{\textcolor[rgb]{0.13,0.47,0.30}{#1}}
\newcommand{\StringTok}[1]{\textcolor[rgb]{0.13,0.47,0.30}{#1}}
\newcommand{\VariableTok}[1]{\textcolor[rgb]{0.07,0.07,0.07}{#1}}
\newcommand{\VerbatimStringTok}[1]{\textcolor[rgb]{0.13,0.47,0.30}{#1}}

\providecommand{\tightlist}{%
  \setlength{\itemsep}{0pt}\setlength{\parskip}{0pt}}\usepackage{longtable,booktabs,array}
\usepackage{calc} 
\usepackage{etoolbox}
\makeatletter
\patchcmd\longtable{\par}{\if@noskipsec\mbox{}\fi\par}{}{}
\makeatother
\IfFileExists{footnotehyper.sty}{\usepackage{footnotehyper}}{\usepackage{footnote}}
\makesavenoteenv{longtable}
\usepackage{graphicx}
\makeatletter
\newsavebox\pandoc@box
\newcommand*\pandocbounded[1]{
  \sbox\pandoc@box{#1}%
  \Gscale@div\@tempa{\textheight}{\dimexpr\ht\pandoc@box+\dp\pandoc@box\relax}%
  \Gscale@div\@tempb{\linewidth}{\wd\pandoc@box}%
  \ifdim\@tempb\p@<\@tempa\p@\let\@tempa\@tempb\fi
  \ifdim\@tempa\p@<\p@\scalebox{\@tempa}{\usebox\pandoc@box}%
  \else\usebox{\pandoc@box}%
  \fi%
}
\def\fps@figure{htbp}
\makeatother
\NewDocumentCommand\citeproctext{}{}
\NewDocumentCommand\citeproc{mm}{%
  \begingroup\def\citeproctext{#2}\cite{#1}\endgroup}
\makeatletter
 \let\@cite@ofmt\@firstofone
 \def\@biblabel#1{}
 \def\@cite#1#2{{#1\if@tempswa , #2\fi}}
\makeatother
\newlength{\cslhangindent}
\newlength{\csllabelwidth}
\newenvironment{CSLReferences}[2] 
 {\begin{list}{}{%
  \setlength{\itemindent}{0pt}
  \setlength{\leftmargin}{0pt}
  \setlength{\parsep}{0pt}
  \ifodd #1
   \setlength{\leftmargin}{\cslhangindent}
   \setlength{\itemindent}{-1\cslhangindent}
  \fi
  \setlength{\itemsep}{#2\baselineskip}}}
 {\end{list}}
\usepackage{calc}

\KOMAoption{captions}{tableheading}
\makeatletter
\@ifpackageloaded{caption}{}{\usepackage{caption}}
\AtBeginDocument{%
\ifdefined\contentsname
  \renewcommand*\contentsname{Table of contents}
\else
  \newcommand\contentsname{Table of contents}
\fi
\ifdefined\listfigurename
  \renewcommand*\listfigurename{List of Figures}
\else
  \newcommand\listfigurename{List of Figures}
\fi
\ifdefined\listtablename
  \renewcommand*\listtablename{List of Tables}
\else
  \newcommand\listtablename{List of Tables}
\fi
\ifdefined\figurename
  \renewcommand*\figurename{Figure}
\else
  \newcommand\figurename{Figure}
\fi
\ifdefined\tablename
  \renewcommand*\tablename{Table}
\else
  \newcommand\tablename{Table}
\fi
}
\@ifpackageloaded{float}{}{\usepackage{float}}
\floatstyle{ruled}
\@ifundefined{c@chapter}{\newfloat{codelisting}{h}{lop}}{\newfloat{codelisting}{h}{lop}[chapter]}
\floatname{codelisting}{Listing}

\makeatother
\makeatletter
\@ifpackageloaded{caption}{}{\usepackage{caption}}
\@ifpackageloaded{subcaption}{}{\usepackage{subcaption}}
\makeatother

\usepackage{bookmark}

\IfFileExists{xurl.sty}{\usepackage{xurl}}{} 
\hypersetup{
  pdftitle={Sentiment Classification of Thai Central Bank Press Releases Using Supervised Learning},
  pdfauthor={Stefano Grassi},
  colorlinks=true,
  linkcolor={blue},
  filecolor={Maroon},
  citecolor={Blue},
  urlcolor={Blue},
  pdfcreator={LaTeX via pandoc}}

\title{Sentiment Classification of Thai Central Bank Press Releases
Using Supervised Learning}
\author{Stefano Grassi}
\date{}

\begin{document}
\maketitle

\section{Introduction}\label{introduction}

In recent decades, central bank communications have not only become a
crucial tool for shaping economic expectations and outcomes. Effective
communication enables central banks to influence financial markets,
enhance monetary policy predictability and achieve macroeconomic goals
(\citeproc{ref-blinder2008}{Blinder et al. 2008}). Research indicates
that policy announcements are pivotal for the implementation of monetary
policy (\citeproc{ref-guthrie2000}{Guthrie and Wright 2000}) and offer
insights into future policy directions
(\citeproc{ref-gurkaynak2005}{Gürkaynak, Sack, and Swanson 2005}). As
Bernanke (\citeproc{ref-bernanke2015}{2015}) noted, communication stands
as one of the most powerful instruments at central banks' disposal,
highlighting the significant risks associated with miscommunication.
This critical role of communication has spurred growing interest in
understanding and analyzing central bank messaging. Coupled with
advancements in Natural Language Processing (NLP) and computational
power, this interest has driven the application of various NLP
techniques to study central bank texts (\citeproc{ref-bholat2015}{Bholat
et al. 2015}). In particular, the literature on sentiment analysis
commonly employs two main approaches: dictionary-based and machine
learning methods (\citeproc{ref-algaba2020}{Algaba et al. 2020}).
Dictionary-based methods rely on predefined sets of n-grams to determine
sentiment based on the frequency of positive and negative terms, with
popular examples including the dictionaries by Loughran and McDonald
(\citeproc{ref-loughran2011}{2011}) and Apel and Grimaldi
(\citeproc{ref-apel2014}{2014}). Conversely, machine learning methods,
particularly supervised learning techniques such as Naive Bayes,
classify sentiment by learning from a pre-labeled corpus and then
applying this knowledge to new, unseen communications. While both
approaches have their strengths and weaknesses, as highlighted by
Frankel et al. (\citeproc{ref-frankel2022}{2022}), the adoption of
supervised learning is less prevalent. This is likely due to the
time-consuming effort and expertise required to construct a pre-labeled
dataset. Additionally, the supervised learning literature is limited in
the variety and size of corpora used, indicating a need for further
empirical evidence across diverse scenarios and central banks,
especially as much of the existing research focuses on developed
economies, leaving emerging markets less represented. To address these
gaps, this study employs supervised machine learning techniques in
Python to classify the sentiment of press releases from the Bank of
Thailand, focusing on the context of Thai monetary policy.

\section{The Dataset}\label{the-dataset}

The corpus comprises 794 pre-labeled sentences extracted from 26
English-language press releases issued by the Bank of Thailand. These
press releases span the period from 3 February 2021 to 18 December 2024.
A representative sample of the dataset is presented in Table 1.

\phantomsection\label{table-1-prelabelled-dataset}
\pandocbounded{\includegraphics[keepaspectratio]{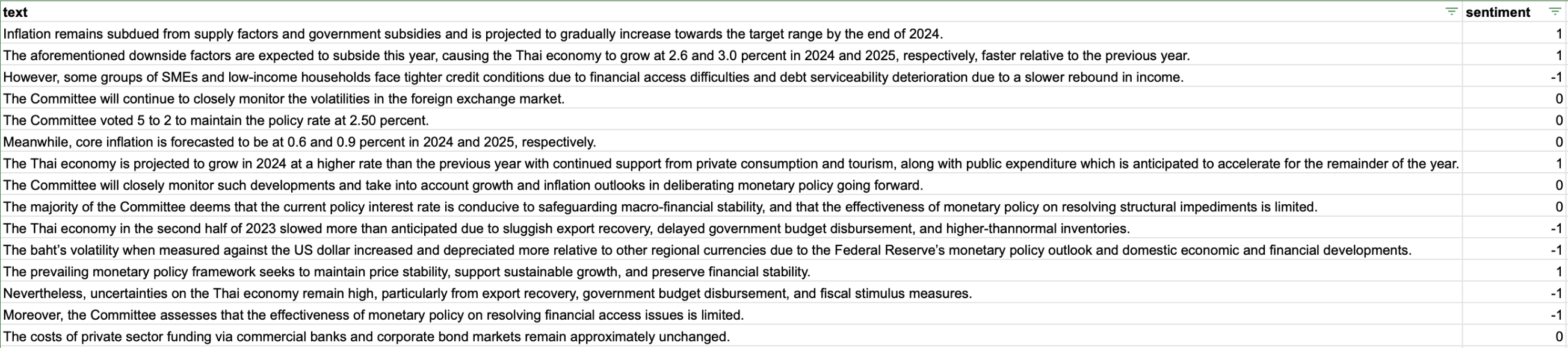}}

Table 1: a sample of pre-labeled sentences from Bank of Thailand press
releases in English.

The dataset was constructed by extracting the main body text from
Monetary Policy Committee (MPC) press releases in PDF format, which are
publicly available on the Thai Central Bank website
(\citeproc{ref-bot_press_statement}{Bank of Thailand, n.d.}). The
extracted text was split into sentences and sentiment-labeled using
ChatGPT, followed by manual revisions by the author to ensure validation
and quality assurance. While this approach combined generative AI and
human expertise, it was time intensive. Additionally, challenges during
annotation, such as sentences with mixed sentiment, required expert
judgment, which introduced bias. While many studies represent sentiment
in a binary fashion, categorizing it as either ``hawkish'' (positive) or
``dovish'' (negative), this approach has been critiqued in the
literature. Although such a classification can be useful in some
contexts, associating policy stance directly with sentiment can be
misleading, especially when negative words are not necessarily
``hawkish'' or vice versa. Therefore, this study opts for a more
straightforward approach in representing sentiment, avoiding
associations with policy stances.

\begin{Shaded}
\begin{Highlighting}[]
\CommentTok{\# Import libraries for loading the dataset}
\ImportTok{import}\NormalTok{ os}
\ImportTok{import}\NormalTok{ re}
\ImportTok{import}\NormalTok{ pandas }\ImportTok{as}\NormalTok{ pd}
\ImportTok{import}\NormalTok{ numpy }\ImportTok{as}\NormalTok{ np}

\CommentTok{\# Load dataset}
\NormalTok{bot\_prelabeled\_sent }\OperatorTok{=}\NormalTok{ pd.read\_csv(}\StringTok{\textquotesingle{}./bot\_annotated\_sentences.csv\textquotesingle{}}\NormalTok{)}

\CommentTok{\# Show info}
\NormalTok{bot\_prelabeled\_sent.info()}
\end{Highlighting}
\end{Shaded}

\begin{verbatim}
<class 'pandas.core.frame.DataFrame'>
RangeIndex: 794 entries, 0 to 793
Data columns (total 2 columns):
 #   Column     Non-Null Count  Dtype 
---  ------     --------------  ----- 
 0   text       794 non-null    object
 1   sentiment  794 non-null    int64 
dtypes: int64(1), object(1)
memory usage: 12.5+ KB
\end{verbatim}

Figure 1 illustrates the distribution of sentiment classes. The minority
class, negative sentiment, accounts for 21.3\% of the total dataset.
Although the corpus is predominantly positive and neutral, the dataset
is not considered highly imbalanced. As such, imbalanced dataset
techniques are not deemed necessary for this study.

\pandocbounded{\includegraphics[keepaspectratio]{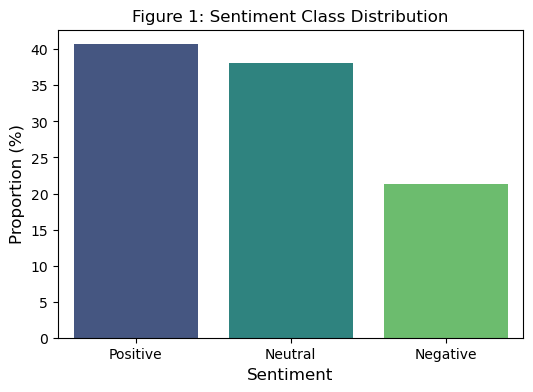}}

Detailed information on the dataset preparation and labeling methodology
can be found in Appendices A and B.

\section{Objective}\label{objective}

The primary objective of this study is to classify the sentiment
(negative, neutral, positive) of English Thai central bank press
releases using three commonly used supervised machine learning
techniques: Naive Bayes (\citeproc{ref-li2010}{Li 2010}), Support Vector
Machines (\citeproc{ref-mullen2004}{Mullen and Collier 2004}) and Random
Forest (\citeproc{ref-frankel2022}{Frankel, Jennings, and Lee 2022}).

This paper contributes to the literature by:

\begin{enumerate}
\def\labelenumi{\arabic{enumi}.}
\tightlist
\item
  \textbf{Expanding the application of machine-learning methods in
  central bank communication}: while existing research predominantly
  relies on lexicon-based methods, few studies explore supervised
  machine learning approaches due to challenges such as the lack of
  labeled datasets. This study aims to address this gap and provide
  robust evidence of the effectiveness of data-driven methods in
  sentiment classification.
\item
  \textbf{Increasing relevance to modern data environments}: in the era
  of Big Data, the adoption of NLP techniques by central banks,
  financial institutions and policymakers is increasingly important.
  These methods not only allow for the automated analysis of large
  volumes of textual data but also offer downstream applications, such
  as improving econometric models and leveraging sentiment features for
  time series analysis. This research highlights the practical benefits
  of integrating sentiment analysis into decision-making processes.
\item
  \textbf{Filling the gap in emerging market research}: much of the
  literature on central bank communication centers on developed
  economies, leaving emerging markets less explored. By focusing on the
  Bank of Thailand, this work offers additional valuable insights into
  the classification of central bank communications in the context of a
  developing economy.
\end{enumerate}

\section{Evaluation Methodology}\label{evaluation-methodology}

The evaluation of the three proposed sentiment classifiers is based on
the following metrics:

\begin{itemize}
\tightlist
\item
  Precision
\item
  Recall
\item
  F1 Score
\item
  Macro-Precision
\item
  Macro-Recall
\item
  Macro-F1 Score
\end{itemize}

These metrics are chosen for two primary reasons. First, precision and
recall are standard in NLP tasks. Precision measures the proportion of
correctly predicted positive instances out of all predicted positives,
while recall assesses the proportion of actual positives correctly
identified. The F1 score, which is the harmonic mean of precision and
recall, balances the trade-off between these metrics. This is
particularly useful when the distribution of sentiment classes is
uneven. Second, the use of these metrics is supported by their
established application in previous research on central bank
communications, such as the study by Pfeifer and Marohl
(\citeproc{ref-pfeifer2023}{2023}). Although their study utilized a
binary classification methodology with a different dataset composition
and size, their findings provide a relevant point of reference for this
research. Since this study involves multinomial sentiment
classification, macroaveraging is employed. Macroaveraging calculates
the metrics independently for each class and then averages them, making
it more sensitive to the performance of smaller classes
(\citeproc{ref-jurafsky2024}{Jurafsky and Martin 2024}). These metrics
are reported both per class and as a macro average. Additionally, the
models are fine-tuned to enhance performance using the Macro-F1 score.
This fine-tuning process, which includes hyperparameter optimization and
validation strategies, is further elaborated in the following sections.
This approach is particularly appropriate when all classes are equally
important, as is the case here, given the imbalanced characteristic of
the dataset.

\section{Preprocessing}\label{preprocessing}

The dataset is first split into 80\% training (635) and 20\% testing
(159) sets in a stratified manner.

\begin{Shaded}
\begin{Highlighting}[]
\CommentTok{\# Import library for splitting}
\ImportTok{from}\NormalTok{ sklearn.model\_selection }\ImportTok{import}\NormalTok{ train\_test\_split}

\CommentTok{\# Prepare Dataset for splitting}
\NormalTok{X }\OperatorTok{=}\NormalTok{ bot\_prelabeled\_sent[}\StringTok{\textquotesingle{}text\textquotesingle{}}\NormalTok{].values }
\NormalTok{y }\OperatorTok{=}\NormalTok{ bot\_prelabeled\_sent[}\StringTok{\textquotesingle{}sentiment\textquotesingle{}}\NormalTok{].values }

\CommentTok{\# Split the data into training and testing sets}
\NormalTok{X\_train, X\_test, y\_train, y\_test }\OperatorTok{=}\NormalTok{ train\_test\_split(}
\NormalTok{    X, y, test\_size}\OperatorTok{=}\FloatTok{0.2}\NormalTok{,}
\NormalTok{    random\_state}\OperatorTok{=}\DecValTok{42}\NormalTok{, }\CommentTok{\# for reproducibility}
\NormalTok{    shuffle}\OperatorTok{=}\VariableTok{True}\NormalTok{,}
\NormalTok{    stratify}\OperatorTok{=}\NormalTok{y }\CommentTok{\# to ensure classes are split appropriately}
\NormalTok{   )}
\CommentTok{\# Print statement}
\BuiltInTok{print}\NormalTok{(}\StringTok{\textquotesingle{}Data is split into 80\% train and 20\% test sets.\textquotesingle{}}\NormalTok{)}
\end{Highlighting}
\end{Shaded}

\begin{verbatim}
Data is split into 80% train and 20% test sets.
\end{verbatim}

Then, the text is preprocessed using a custom tokenizer combined with
the scikit-learn \texttt{TfidfVectorizer} for feature extraction
(\citeproc{ref-scikit-learn}{Pedregosa et al. 2011}), a standard
approach in machine learning methods. The tokenizer captures words,
numbers and hyphenated terms, applies lemmatization using
\texttt{WordNetLemmatizer} from NLTK (\citeproc{ref-bird2009}{Bird,
Loper, and Klein 2009}) to reduce words to their base forms and removes
common English stop words (e.g., ``the'', ``a'') to reduce
dimensionality and speed up computation. The \texttt{TfidfVectorizer}
transforms the text into numerical features using Term Frequency-Inverse
Document Frequency (TF-IDF), which is a numerical measure that indicates
the significance of a word within a document relative to a collection or
corpus (\citeproc{ref-manning2008}{Manning, Raghavan, and Schütze
2008}), capturing the importance of n-grams (unigrams, bigrams and
trigrams) while accounting for term presence/absence and lowercasing to
ensure consistency.

\begin{Shaded}
\begin{Highlighting}[]
\CommentTok{\# Import libraries for preprocessing}
\ImportTok{from}\NormalTok{ sklearn.feature\_extraction.text }\ImportTok{import}\NormalTok{ TfidfVectorizer}
\ImportTok{from}\NormalTok{ sklearn.feature\_extraction.text }\ImportTok{import}\NormalTok{ ENGLISH\_STOP\_WORDS}
\ImportTok{from}\NormalTok{ nltk.stem }\ImportTok{import}\NormalTok{ WordNetLemmatizer}


\CommentTok{\# Initialize the lemmatizer}
\NormalTok{lemmatizer }\OperatorTok{=}\NormalTok{ WordNetLemmatizer()}

\CommentTok{\# Define a custom tokenizer}
\KeywordTok{def}\NormalTok{ custom\_tokenizer(text):}
    \CommentTok{"""}
\CommentTok{    Tokenizes the input text by extracting words, numbers, }
\CommentTok{    hyphenated terms and applying lemmatization to reduce }
\CommentTok{    words to their base forms.}
\CommentTok{    }
\CommentTok{    Args:}
\CommentTok{    text (str): The input text to be tokenized.}

\CommentTok{    Returns:}
\CommentTok{    list: A list of lemmatized tokens after removing common }
\CommentTok{    English stop words.}
\CommentTok{    }
\CommentTok{    The function uses regular expressions to capture words, }
\CommentTok{    floating{-}point numbers and integers, then applies }
\CommentTok{    lemmatization using WordNetLemmatizer from NLTK.}
\CommentTok{    It also removes stop words to reduce dimensionality }
\CommentTok{    and improve computation speed.}
\CommentTok{    """}
    \CommentTok{\# Capture words, floating{-}point numbers and integers}
\NormalTok{    tokens }\OperatorTok{=}\NormalTok{ re.findall(}\VerbatimStringTok{r\textquotesingle{}\textbackslash{}b\textbackslash{}d+\textbackslash{}.\textbackslash{}d+|\textbackslash{}b\textbackslash{}d+|\textbackslash{}b\textbackslash{}w+(?:{-}\textbackslash{}w+)?\textbackslash{}b\textquotesingle{}}\NormalTok{, text)}
    
    \CommentTok{\# Remove stop words and apply lemmatization}
    \ControlFlowTok{return}\NormalTok{ [}
\NormalTok{        lemmatizer.lemmatize(token) }
        \ControlFlowTok{for}\NormalTok{ token }\KeywordTok{in}\NormalTok{ tokens }
        \ControlFlowTok{if}\NormalTok{ token.lower() }\KeywordTok{not} \KeywordTok{in}\NormalTok{ ENGLISH\_STOP\_WORDS}
\NormalTok{    ]}

\CommentTok{\# Apply TF{-}IDF vectorization}
\NormalTok{tf\_idf }\OperatorTok{=}\NormalTok{ TfidfVectorizer(}
\NormalTok{    ngram\_range}\OperatorTok{=}\NormalTok{(}\DecValTok{1}\NormalTok{, }\DecValTok{3}\NormalTok{),  }\CommentTok{\# Use unigrams, bigrams and trigrams}
\NormalTok{    binary}\OperatorTok{=}\VariableTok{True}\NormalTok{,         }\CommentTok{\# presence/absence of words}
\NormalTok{    smooth\_idf}\OperatorTok{=}\VariableTok{False}\NormalTok{,    }\CommentTok{\# Disable smoothing of idf}
\NormalTok{    lowercase}\OperatorTok{=}\VariableTok{True}\NormalTok{,      }\CommentTok{\# Ensure all text is converted to lowercase}
\NormalTok{    token\_pattern}\OperatorTok{=}\VariableTok{None}\NormalTok{,  }\CommentTok{\# Explicitly allow custom tokenizer}
\NormalTok{    tokenizer}\OperatorTok{=}\NormalTok{custom\_tokenizer  }\CommentTok{\# Use custom tokenizer}
\NormalTok{)}

\CommentTok{\# Fit and transform the training data, and transform the testing data}
\NormalTok{X\_train\_tfidf }\OperatorTok{=}\NormalTok{ tf\_idf.fit\_transform(X\_train)}
\NormalTok{X\_test\_tfidf }\OperatorTok{=}\NormalTok{ tf\_idf.transform(X\_test)}
\end{Highlighting}
\end{Shaded}

\subsection{Baseline Performance}\label{baseline-performance}

Among the three classifiers selected for this study, Naive Bayes serves
as the baseline. It is a widely used method in NLP sentiment
classification tasks due to its simplicity, making it a natural choice
for comparison.

\section{Classification Approach}\label{classification-approach}

The classifiers are implemented using the scikit-learn library and are
fine-tuned to optimize the Macro-F1 score through grid search with
5-fold cross-validation (\citeproc{ref-james2013}{James et al. 2013}).
The selection of the hyperparameters takes was based on several tests
run on a Mac M1 hardware.

\subsection{Naive Bayes}\label{naive-bayes}

Naive Bayes has a single hyperparameter, \texttt{alpha}, which is used
for Laplace smoothing to address the issue of zero probability when a
new observation appears in the query data
(\citeproc{ref-analyticsvidhya2021}{Analytics Vidhya 2021}).

\begin{Shaded}
\begin{Highlighting}[]
\CommentTok{\# Import Grid Search CV }
\ImportTok{from}\NormalTok{ sklearn.model\_selection }\ImportTok{import}\NormalTok{ GridSearchCV}

\CommentTok{\# Import NB model}
\ImportTok{from}\NormalTok{ sklearn.naive\_bayes }\ImportTok{import}\NormalTok{ MultinomialNB}

\CommentTok{\# Define the model}
\NormalTok{nb\_model }\OperatorTok{=}\NormalTok{ MultinomialNB()}

\CommentTok{\# Define the parameter grid for alpha values}
\NormalTok{param\_grid }\OperatorTok{=}\NormalTok{ \{}\StringTok{\textquotesingle{}alpha\textquotesingle{}}\NormalTok{: [}\FloatTok{0.00001}\NormalTok{, }\FloatTok{0.0001}\NormalTok{, }\FloatTok{0.001}\NormalTok{, }\FloatTok{0.1}\NormalTok{, }\DecValTok{1}\NormalTok{, }\DecValTok{10}\NormalTok{, }\DecValTok{100}\NormalTok{, }\DecValTok{1000}\NormalTok{]\}}

\CommentTok{\# Perform GridSearchCV with 5{-}fold cross{-}validation and macro F1 score}
\NormalTok{grid\_search }\OperatorTok{=}\NormalTok{ GridSearchCV(nb\_model, }
\NormalTok{                           param\_grid, }
\NormalTok{                           scoring}\OperatorTok{=}\StringTok{\textquotesingle{}f1\_macro\textquotesingle{}}\NormalTok{, cv}\OperatorTok{=}\DecValTok{5}\NormalTok{)}

\CommentTok{\# Fit the Naive Bayes Model model}
\NormalTok{grid\_search.fit(X\_train\_tfidf, y\_train)}

\CommentTok{\# Best hyperparameter value for alpha}
\NormalTok{best\_alpha }\OperatorTok{=}\NormalTok{ grid\_search.best\_params\_[}\StringTok{\textquotesingle{}alpha\textquotesingle{}}\NormalTok{]}
\BuiltInTok{print}\NormalTok{(}\SpecialStringTok{f"Best alpha (based on Macro{-}F1 score): }\SpecialCharTok{\{}\NormalTok{best\_alpha}\SpecialCharTok{\}}\SpecialStringTok{"}\NormalTok{)}

\CommentTok{\# Fit the NB model}
\NormalTok{best\_nb\_model }\OperatorTok{=}\NormalTok{ MultinomialNB(alpha}\OperatorTok{=}\NormalTok{best\_alpha)}
\NormalTok{best\_nb\_model }\OperatorTok{=}\NormalTok{ best\_nb\_model.fit(X\_train\_tfidf, y\_train)}
\end{Highlighting}
\end{Shaded}

\begin{verbatim}
Best alpha (based on Macro-F1 score): 0.1
\end{verbatim}

\subsection{Support Vector Machines}\label{support-vector-machines}

The key hyperparameters of Support Vector Machines (SVMs) are \texttt{C}
(regularization parameter), \texttt{gamma} (kernel coefficient,
influencing the decision boundary) and \texttt{kernel} (type of function
to map data, such as \texttt{linear}, \texttt{rbf} or \texttt{poly}).

\begin{Shaded}
\begin{Highlighting}[]
\CommentTok{\# Import SVM model and related libraries}
\ImportTok{from}\NormalTok{ sklearn }\ImportTok{import}\NormalTok{ svm, model\_selection}

\CommentTok{\# Define SVM classifier and parameters}
\NormalTok{svm\_clf }\OperatorTok{=}\NormalTok{ svm.SVC(probability}\OperatorTok{=}\VariableTok{True}\NormalTok{)}

\CommentTok{\# Hyperparameter tuning with GridSearchCV, }
\CommentTok{\# scoring based on macro{-}F1 score}
\NormalTok{parameters }\OperatorTok{=}\NormalTok{ \{}
    \StringTok{\textquotesingle{}C\textquotesingle{}}\NormalTok{: [}\FloatTok{0.1}\NormalTok{, }\DecValTok{1}\NormalTok{, }\DecValTok{10}\NormalTok{], }\CommentTok{\# regularization parameter}
    \StringTok{\textquotesingle{}kernel\textquotesingle{}}\NormalTok{: [}\StringTok{\textquotesingle{}linear\textquotesingle{}}\NormalTok{, }\StringTok{\textquotesingle{}rbf\textquotesingle{}}\NormalTok{, }\StringTok{\textquotesingle{}poly\textquotesingle{}}\NormalTok{], }\CommentTok{\# type of function to map data}
    \StringTok{\textquotesingle{}gamma\textquotesingle{}}\NormalTok{: [}\FloatTok{0.01}\NormalTok{, }\FloatTok{0.1}\NormalTok{, }\FloatTok{0.5}\NormalTok{] }\CommentTok{\# kernel coefficient}
\NormalTok{\} }

\NormalTok{grid\_svm\_clf }\OperatorTok{=}\NormalTok{ model\_selection.GridSearchCV(estimator}\OperatorTok{=}\NormalTok{svm\_clf, }
\NormalTok{                                            param\_grid}\OperatorTok{=}\NormalTok{parameters, }
\NormalTok{                                            scoring}\OperatorTok{=}\StringTok{\textquotesingle{}f1\_macro\textquotesingle{}}\NormalTok{, cv}\OperatorTok{=}\DecValTok{5}\NormalTok{)}

\CommentTok{\# Fit the grid search to the training data}
\NormalTok{grid\_svm\_clf.fit(X\_train\_tfidf, y\_train)}

\CommentTok{\# Best hyperparameter combination based on macro F1 score}
\NormalTok{best\_params }\OperatorTok{=}\NormalTok{ grid\_svm\_clf.best\_params\_}
\BuiltInTok{print}\NormalTok{(}\SpecialStringTok{f"Best parameters (based on Macro{-}F1 score):}\CharTok{\textbackslash{}n}\SpecialCharTok{\{}\NormalTok{best\_params}\SpecialCharTok{\}}\SpecialStringTok{"}\NormalTok{)}

\CommentTok{\# Get Best SVM Model}
\NormalTok{best\_svm\_model }\OperatorTok{=}\NormalTok{ grid\_svm\_clf.best\_estimator\_}
\end{Highlighting}
\end{Shaded}

\begin{verbatim}
Best parameters (based on Macro-F1 score):
{'C': 10, 'gamma': 0.01, 'kernel': 'linear'}
\end{verbatim}

\subsection{Random Forest}\label{random-forest}

The key hyperparameters of Random Forest are \texttt{n\_estimators}
(number of trees), \texttt{max\_depth} (maximum depth of trees),
\texttt{min\_samples\_split} (minimum samples required to split a node)
and \texttt{max\_features} (maximum number of features to consider for
splitting).

\begin{Shaded}
\begin{Highlighting}[]
\CommentTok{\# Import Random Forest}
\ImportTok{from}\NormalTok{ sklearn.ensemble }\ImportTok{import}\NormalTok{ RandomForestClassifier}

\CommentTok{\# Define the Random Forest classifier}
\NormalTok{rf\_clf }\OperatorTok{=}\NormalTok{ RandomForestClassifier(random\_state}\OperatorTok{=}\DecValTok{42}\NormalTok{)}

\CommentTok{\# Hyperparameter grid with a reduced search space}
\NormalTok{parameters }\OperatorTok{=}\NormalTok{ \{}
    \StringTok{\textquotesingle{}n\_estimators\textquotesingle{}}\NormalTok{: [}\DecValTok{100}\NormalTok{, }\DecValTok{200}\NormalTok{],}
    \StringTok{\textquotesingle{}max\_depth\textquotesingle{}}\NormalTok{: [}\DecValTok{20}\NormalTok{, }\DecValTok{30}\NormalTok{],}
    \StringTok{\textquotesingle{}min\_samples\_split\textquotesingle{}}\NormalTok{: [}\DecValTok{2}\NormalTok{, }\DecValTok{5}\NormalTok{],}
    \StringTok{\textquotesingle{}max\_features\textquotesingle{}}\NormalTok{: [}\StringTok{\textquotesingle{}sqrt\textquotesingle{}}\NormalTok{, }\StringTok{\textquotesingle{}log2\textquotesingle{}}\NormalTok{],}
\NormalTok{\}}

\CommentTok{\# Apply GridSearchCV }
\NormalTok{grid\_rf\_clf }\OperatorTok{=}\NormalTok{ GridSearchCV(rf\_clf, param\_grid}\OperatorTok{=}\NormalTok{parameters, cv}\OperatorTok{=}\DecValTok{5}\NormalTok{, }
\NormalTok{                           scoring}\OperatorTok{=}\StringTok{\textquotesingle{}f1\_macro\textquotesingle{}}\NormalTok{, n\_jobs}\OperatorTok{={-}}\DecValTok{1}\NormalTok{)}

\CommentTok{\# Fit the grid search }
\NormalTok{grid\_rf\_clf.fit(X\_train\_tfidf, y\_train)}

\CommentTok{\# Get the best parameters}
\NormalTok{best\_params }\OperatorTok{=}\NormalTok{ grid\_rf\_clf.best\_params\_}

\CommentTok{\# Print the best parameters}
\BuiltInTok{print}\NormalTok{(}\StringTok{"Best parameters (based on Macro{-}F1 score):}\CharTok{\textbackslash{}n}\StringTok{"}
      \SpecialStringTok{f"max\_depth: }\SpecialCharTok{\{}\NormalTok{best\_params[}\StringTok{\textquotesingle{}max\_depth\textquotesingle{}}\NormalTok{]}\SpecialCharTok{\}}\SpecialStringTok{,}\CharTok{\textbackslash{}n}\SpecialStringTok{"}
      \SpecialStringTok{f"max\_features: }\SpecialCharTok{\{}\NormalTok{best\_params[}\StringTok{\textquotesingle{}max\_features\textquotesingle{}}\NormalTok{]}\SpecialCharTok{\}}\SpecialStringTok{,}\CharTok{\textbackslash{}n}\SpecialStringTok{"}
      \SpecialStringTok{f"min\_samples\_split: }\SpecialCharTok{\{}\NormalTok{best\_params[}\StringTok{\textquotesingle{}min\_samples\_split\textquotesingle{}}\NormalTok{]}\SpecialCharTok{\}}\SpecialStringTok{,}\CharTok{\textbackslash{}n}\SpecialStringTok{"}
      \SpecialStringTok{f"n\_estimators: }\SpecialCharTok{\{}\NormalTok{best\_params[}\StringTok{\textquotesingle{}n\_estimators\textquotesingle{}}\NormalTok{]}\SpecialCharTok{\}}\SpecialStringTok{"}\NormalTok{)}

\CommentTok{\# Get best RF Model}
\NormalTok{best\_rf\_model }\OperatorTok{=}\NormalTok{ grid\_rf\_clf.best\_estimator\_}
\end{Highlighting}
\end{Shaded}

\begin{verbatim}
Best parameters (based on Macro-F1 score):
max_depth: 30,
max_features: sqrt,
min_samples_split: 2,
n_estimators: 100
\end{verbatim}

\section{Evaluation}\label{evaluation}

The performance of the three classifiers summarised in Table 2, reveals
insights in concordance with the literature on sentiment classification.
Naive Bayes outperforms both Random Forest and SVM with a Macro-F1 score
of 0.75, compared to SVM's 0.71 and Random Forest's 0.62. This indicates
that NB provides the most balanced performance across all classes.
Notably, the relatively small sample size likely contributes to these
results, as more data would enable the more complex models like Random
Forest and SVM to outperform Naive Bayes, as seen in other studies.

\begin{Shaded}
\begin{Highlighting}[]
\CommentTok{\# Import Classification Report from sklearn}
\ImportTok{from}\NormalTok{ sklearn.metrics }\ImportTok{import}\NormalTok{ classification\_report}

\CommentTok{\# Define the target names (sentiment classes)}
\NormalTok{target\_names }\OperatorTok{=}\NormalTok{ [}\StringTok{\textquotesingle{}Negative\textquotesingle{}}\NormalTok{, }\StringTok{\textquotesingle{}Neutral\textquotesingle{}}\NormalTok{, }\StringTok{\textquotesingle{}Positive\textquotesingle{}}\NormalTok{]}

\CommentTok{\# Get the predictions for the test set}
\NormalTok{y\_pred\_rf }\OperatorTok{=}\NormalTok{ best\_rf\_model.predict(X\_test\_tfidf)}
\NormalTok{y\_pred\_svm }\OperatorTok{=}\NormalTok{ best\_svm\_model.predict(X\_test\_tfidf)}
\NormalTok{y\_pred\_nb }\OperatorTok{=}\NormalTok{ best\_nb\_model.predict(X\_test\_tfidf)}

\CommentTok{\# Get classification report for each model}
\NormalTok{report\_rf }\OperatorTok{=}\NormalTok{ classification\_report(y\_test, y\_pred\_rf, }
\NormalTok{                                  target\_names}\OperatorTok{=}\NormalTok{target\_names, }
\NormalTok{                                  output\_dict}\OperatorTok{=}\VariableTok{True}\NormalTok{)}
\NormalTok{report\_svm }\OperatorTok{=}\NormalTok{ classification\_report(y\_test, y\_pred\_svm, }
\NormalTok{                                   target\_names}\OperatorTok{=}\NormalTok{target\_names, }
\NormalTok{                                   output\_dict}\OperatorTok{=}\VariableTok{True}\NormalTok{)}
\NormalTok{report\_nb }\OperatorTok{=}\NormalTok{ classification\_report(y\_test, y\_pred\_nb, }
\NormalTok{                                  target\_names}\OperatorTok{=}\NormalTok{target\_names, }
\NormalTok{                                  output\_dict}\OperatorTok{=}\VariableTok{True}\NormalTok{)}

\CommentTok{\# Create DataFrames for class{-}wise metrics}
\NormalTok{df\_rf }\OperatorTok{=}\NormalTok{ pd.DataFrame(report\_rf).transpose()}
\NormalTok{df\_svm }\OperatorTok{=}\NormalTok{ pd.DataFrame(report\_svm).transpose()}
\NormalTok{df\_nb }\OperatorTok{=}\NormalTok{ pd.DataFrame(report\_nb).transpose()}

\CommentTok{\# Select metrics to report for each classifier}
\NormalTok{df\_rf }\OperatorTok{=}\NormalTok{ df\_rf[[}\StringTok{\textquotesingle{}precision\textquotesingle{}}\NormalTok{, }
               \StringTok{\textquotesingle{}recall\textquotesingle{}}\NormalTok{, }
               \StringTok{\textquotesingle{}f1{-}score\textquotesingle{}}\NormalTok{]].drop([}\StringTok{\textquotesingle{}accuracy\textquotesingle{}}\NormalTok{, }
                                  \StringTok{\textquotesingle{}weighted avg\textquotesingle{}}\NormalTok{]).}\BuiltInTok{round}\NormalTok{(}\DecValTok{2}\NormalTok{)}
\NormalTok{df\_svm }\OperatorTok{=}\NormalTok{ df\_svm[[}\StringTok{\textquotesingle{}precision\textquotesingle{}}\NormalTok{, }
                 \StringTok{\textquotesingle{}recall\textquotesingle{}}\NormalTok{, }
                 \StringTok{\textquotesingle{}f1{-}score\textquotesingle{}}\NormalTok{]].drop([}\StringTok{\textquotesingle{}accuracy\textquotesingle{}}\NormalTok{, }
                                    \StringTok{\textquotesingle{}weighted avg\textquotesingle{}}\NormalTok{]).}\BuiltInTok{round}\NormalTok{(}\DecValTok{2}\NormalTok{)}
\NormalTok{df\_nb }\OperatorTok{=}\NormalTok{ df\_nb[[}\StringTok{\textquotesingle{}precision\textquotesingle{}}\NormalTok{, }
               \StringTok{\textquotesingle{}recall\textquotesingle{}}\NormalTok{, }
               \StringTok{\textquotesingle{}f1{-}score\textquotesingle{}}\NormalTok{]].drop([}\StringTok{\textquotesingle{}accuracy\textquotesingle{}}\NormalTok{, }
                                  \StringTok{\textquotesingle{}weighted avg\textquotesingle{}}\NormalTok{]).}\BuiltInTok{round}\NormalTok{(}\DecValTok{2}\NormalTok{)}


\CommentTok{\# List of dataframes and corresponding model names}
\NormalTok{models\_df }\OperatorTok{=}\NormalTok{ [df\_rf, df\_svm, df\_nb]}

\CommentTok{\# Initialize an empty list to store the final result}
\NormalTok{final\_data }\OperatorTok{=}\NormalTok{ []}

\CommentTok{\# Loop through each model\textquotesingle{}s dataframe}
\ControlFlowTok{for}\NormalTok{ df }\KeywordTok{in}\NormalTok{ models\_df:}
    \CommentTok{\# Add a header row for the model name spanning columns}
\NormalTok{    header\_row }\OperatorTok{=}\NormalTok{ pd.DataFrame([[}\StringTok{\textquotesingle{}\textquotesingle{}}\NormalTok{, }\StringTok{\textquotesingle{}\textquotesingle{}}\NormalTok{, }\StringTok{\textquotesingle{}\textquotesingle{}}\NormalTok{]], columns}\OperatorTok{=}\NormalTok{df.columns)}
    
    \CommentTok{\# Concatenate: header row and the dataframe}
\NormalTok{    final\_data.append(header\_row)}
\NormalTok{    final\_data.append(df.reset\_index())}

\CommentTok{\# Concatenate all dataframes}
\NormalTok{final\_df }\OperatorTok{=}\NormalTok{ pd.concat(final\_data, ignore\_index}\OperatorTok{=}\VariableTok{True}\NormalTok{)}

\CommentTok{\# Manually assign model names to specific rows using iloc}
\NormalTok{final\_df.iloc[}\DecValTok{0}\NormalTok{, }\DecValTok{3}\NormalTok{] }\OperatorTok{=} \StringTok{\textquotesingle{}Random Forest\textquotesingle{}}  
\NormalTok{final\_df.iloc[}\DecValTok{5}\NormalTok{, }\DecValTok{3}\NormalTok{] }\OperatorTok{=} \StringTok{\textquotesingle{}SVM\textquotesingle{}}           
\NormalTok{final\_df.iloc[}\DecValTok{10}\NormalTok{, }\DecValTok{3}\NormalTok{] }\OperatorTok{=} \StringTok{\textquotesingle{}Naive Bayes\textquotesingle{}}  

\CommentTok{\# Reorder columns to put \textquotesingle{}index\textquotesingle{} first}
\NormalTok{final\_df }\OperatorTok{=}\NormalTok{ final\_df[[}\StringTok{\textquotesingle{}index\textquotesingle{}}\NormalTok{] }\OperatorTok{+} 
\NormalTok{                     [col }\ControlFlowTok{for}\NormalTok{ col }\KeywordTok{in}\NormalTok{ final\_df.columns }\ControlFlowTok{if}\NormalTok{ col }\OperatorTok{!=} \StringTok{\textquotesingle{}index\textquotesingle{}}\NormalTok{]]}


\CommentTok{\# Fill NaN values with empty strings for markdown compatibility}
\NormalTok{final\_df.fillna(}\StringTok{\textquotesingle{}\textquotesingle{}}\NormalTok{, inplace}\OperatorTok{=}\VariableTok{True}\NormalTok{)}

\CommentTok{\# Rename \textquotesingle{}index\textquotesingle{} column to empty for markdown}
\NormalTok{final\_df.rename(columns}\OperatorTok{=}\NormalTok{\{}\StringTok{\textquotesingle{}index\textquotesingle{}}\NormalTok{: }\StringTok{\textquotesingle{}\textquotesingle{}}\NormalTok{\}, inplace}\OperatorTok{=}\VariableTok{True}\NormalTok{)}
\end{Highlighting}
\end{Shaded}

\textbf{Table 2: Comparison of Model Performance Metrics}

\begin{longtable}[]{@{}llll@{}}
\toprule\noalign{}
& precision & recall & f1-score \\
\midrule\noalign{}
\endhead
\bottomrule\noalign{}
\endlastfoot
Random Forest & & & \\
Negative & 0.65 & 0.44 & 0.53 \\
Neutral & 0.68 & 0.63 & 0.66 \\
Positive & 0.61 & 0.75 & 0.68 \\
macro avg & 0.65 & 0.61 & 0.62 \\
SVM & & & \\
Negative & 0.71 & 0.71 & 0.71 \\
Neutral & 0.66 & 0.68 & 0.67 \\
Positive & 0.78 & 0.75 & 0.77 \\
macro avg & 0.71 & 0.71 & 0.71 \\
Naive Bayes & & & \\
Negative & 0.69 & 0.79 & 0.74 \\
Neutral & 0.75 & 0.68 & 0.71 \\
Positive & 0.78 & 0.78 & 0.78 \\
macro avg & 0.74 & 0.75 & 0.75 \\
\end{longtable}

\section{Conclusion}\label{conclusion}

The primary objective of this study is to classify the sentiment of
central bank communications using the English press releases of the Bank
of Thailand and supervised learning techniques. These techniques
demonstrates their value in automating sentiment analysis, capturing
linguistic nuances and providing more objective metrics and baselines to
counterbalance the limitations of dictionary-based methods. Furthermore,
the emergence of advanced NLP models, such as large language models
(LLMs), suggests the potential for richer interpretations of central
bank communications (\citeproc{ref-alonso2023}{Alonso-Robisco and Carbó
2023}; \citeproc{ref-pfeifer2023}{Pfeifer and Marohl 2023}). In the
context of Thailand, future research could focus on expanding the
English-language corpus to include additional press releases, thereby
enhancing classifier performance. Another avenue of research could
explore less-studied machine learning methods, address issues related to
imbalanced data and contribute to the refinement of labeling
methodologies, building upon the approach outlined in Appendix B.
Additionally, research could focus on classifying press releases in
Thai, addressing language-specific challenges and applying machine
learning techniques to develop an effective sentiment classification
model for Thai texts. Despite these advancements, several operational
challenges were encountered, including dataset acquisition,
preprocessing and labeling. While generative AI and the relatively small
size of the corpus facilitated some steps, the process remained
time-intensive and required human expertise, introducing the risk of
bias, aligning with existing claims. To support research efforts, this
paper recommends that central banks, such as the Bank of Thailand,
enhance accessibility by providing APIs for retrieving textual data and
maintaining comprehensive archives of past communications. In
conclusion, supervised learning approaches, including simple models like
Naive Bayes, offer a promising starting point for sentiment analysis,
even with limited data. By overcoming the challenge of constructing a
dataset, they provide a foundation for exploring more sophisticated
models as the field continues to evolve, effectively complementing
dictionary-based methods and underline the value of machine learning
techniques for quantifying sentiment in central bank communications
within today's data-driven era.

\section{References}\label{references}

\phantomsection\label{refs}

\section{Appendix A: Extracting and Cleaning Text from MPC Press
Releases
PDFs}\label{appendix-a-extracting-and-cleaning-text-from-mpc-press-releases-pdfs}

The process of extracting and cleaning raw text from PDF files is
implemented to ensure the data is well-structured and free from
irrelevant content. Text is extracted using the pdfminer library, and
subsequently cleaned through a series of steps. This includes removing
unwanted elements such as emails, phone numbers, headers, footers and
decision identifiers. Additional refinements address issues like
hyphenation across line breaks, redundant spaces, and page numbers. The
cleaned text is structured into a DataFrame, with each document assigned
a unique identifier for further analysis.

\begin{Shaded}
\begin{Highlighting}[]
\CommentTok{\#\# Import function used for extracting text from MPC press releases PDFs}
\ImportTok{from}\NormalTok{ pdfminer.high\_level }\ImportTok{import}\NormalTok{ extract\_text}

\CommentTok{\# Step 1: function to extract text from a PDF file}
\KeywordTok{def}\NormalTok{ extract\_text\_from\_pdf(pdf\_path):}
    \CommentTok{"""}
\CommentTok{    Extracts text from a PDF file using pdfminer.}
\CommentTok{    """}
    \ControlFlowTok{try}\NormalTok{:}
        \ControlFlowTok{return}\NormalTok{ extract\_text(pdf\_path)}
    \ControlFlowTok{except} \PreprocessorTok{Exception} \ImportTok{as}\NormalTok{ e:}
        \BuiltInTok{print}\NormalTok{(}\SpecialStringTok{f"Error extracting text from }\SpecialCharTok{\{}\NormalTok{pdf\_path}\SpecialCharTok{\}}\SpecialStringTok{: }\SpecialCharTok{\{}\NormalTok{e}\SpecialCharTok{\}}\SpecialStringTok{"}\NormalTok{)}
        \ControlFlowTok{return} \VariableTok{None}

\CommentTok{\# Step 2: function to clean the extracted text}
\KeywordTok{def}\NormalTok{ clean\_extracted\_text(text):}
    \CommentTok{"""}
\CommentTok{    Cleans the extracted text by removing unwanted content, }
\CommentTok{    such as emails, phone numbers,Bank of Thailand footer, }
\CommentTok{    Press Conference, reference numbers, decision identifiers}
\CommentTok{    and extra spaces.}
\CommentTok{    """}
    \CommentTok{\# Remove \textquotesingle{}Press Conference\textquotesingle{} section}
\NormalTok{    text }\OperatorTok{=}\NormalTok{ re.sub(}\VerbatimStringTok{r\textquotesingle{}Press Conference.*\textquotesingle{}}\NormalTok{, }\StringTok{\textquotesingle{}\textquotesingle{}}\NormalTok{, text, flags}\OperatorTok{=}\NormalTok{re.DOTALL)}

    \CommentTok{\# Remove contact information (emails and phone numbers)}
\NormalTok{    text }\OperatorTok{=}\NormalTok{ re.sub(}\VerbatimStringTok{r\textquotesingle{}\textbackslash{}S+@\textbackslash{}S+\textquotesingle{}}\NormalTok{, }\StringTok{\textquotesingle{}\textquotesingle{}}\NormalTok{, text)  }\CommentTok{\# Remove emails}
\NormalTok{    text }\OperatorTok{=}\NormalTok{ re.sub(}\VerbatimStringTok{r\textquotesingle{}\textbackslash{}+?\textbackslash{}d[\textbackslash{}d {-}]\{8,\}\textbackslash{}d\textquotesingle{}}\NormalTok{, }\StringTok{\textquotesingle{}\textquotesingle{}}\NormalTok{, text) }\CommentTok{\# Remove phone numbers}

    \CommentTok{\# Remove specific sections related to Bank of Thailand }
    \CommentTok{\# and incomplete footers}
\NormalTok{    footer\_pattern }\OperatorTok{=}\NormalTok{ (}
     \VerbatimStringTok{r\textquotesingle{}(Bank of Thailand\textbackslash{}s*\textbackslash{}d\{1,2\}\textbackslash{}s*\textbackslash{}w+\textbackslash{}s*\textbackslash{}d}\SpecialCharTok{\{4\}}\VerbatimStringTok{)\textquotesingle{}}
     \VerbatimStringTok{r\textquotesingle{}(.*?(Tel.*?|E{-}?mail:.*?|Monetary Policy Strategy Division.*?))?\textquotesingle{}}
     \VerbatimStringTok{r\textquotesingle{}[\textbackslash{}s]*\textquotesingle{}}
\NormalTok{    )}
\NormalTok{    text }\OperatorTok{=}\NormalTok{ re.sub(footer\_pattern, }\StringTok{\textquotesingle{}\textquotesingle{}}\NormalTok{, }
\NormalTok{                  text, flags}\OperatorTok{=}\NormalTok{re.IGNORECASE }\OperatorTok{|}\NormalTok{ re.DOTALL)}

    \CommentTok{\# Specifically remove incomplete footer lines }
    \CommentTok{\# and cases like ": , E{-}mail:" and "Tel"}
\NormalTok{    text }\OperatorTok{=}\NormalTok{ re.sub(}\VerbatimStringTok{r\textquotesingle{}:\textbackslash{}s*,?\textbackslash{}s*E{-}?mail\textbackslash{}s*:?,?\textbackslash{}s*\textquotesingle{}}\NormalTok{, }\StringTok{\textquotesingle{}\textquotesingle{}}\NormalTok{, }
\NormalTok{                  text, flags}\OperatorTok{=}\NormalTok{re.IGNORECASE)}
    \CommentTok{\# Remove "Tel" with optional punctuation}
\NormalTok{    text }\OperatorTok{=}\NormalTok{ re.sub(}\VerbatimStringTok{r\textquotesingle{}\textbackslash{}bTel[:\textbackslash{}s,]*\textquotesingle{}}\NormalTok{, }\StringTok{\textquotesingle{}\textquotesingle{}}\NormalTok{, }
\NormalTok{                  text, flags}\OperatorTok{=}\NormalTok{re.IGNORECASE) }

    \CommentTok{\# Remove headings and identifiers}
\NormalTok{    text }\OperatorTok{=}\NormalTok{ re.sub(}\VerbatimStringTok{r\textquotesingle{}No\textbackslash{}.\textbackslash{}s*\textbackslash{}d+/\textbackslash{}d+\textquotesingle{}}\NormalTok{, }\StringTok{\textquotesingle{}\textquotesingle{}}\NormalTok{,}
\NormalTok{                  text)  }\CommentTok{\# Remove reference numbers}
    \CommentTok{\# Remove decision identifiers}
\NormalTok{    text }\OperatorTok{=}\NormalTok{ re.sub(}\VerbatimStringTok{r\textquotesingle{}Monetary Policy Committee’s Decision \textbackslash{}d+/\textbackslash{}d+\textquotesingle{}}\NormalTok{, }
                  \StringTok{\textquotesingle{}\textquotesingle{}}\NormalTok{, text)}
\NormalTok{    text }\OperatorTok{=}\NormalTok{ re.sub(}\VerbatimStringTok{r\textquotesingle{}Mr\textbackslash{}.\textbackslash{}s+\textbackslash{}w+\textquotesingle{}}\NormalTok{, }\StringTok{\textquotesingle{}\textquotesingle{}}\NormalTok{, text)  }\CommentTok{\# Remove names}

    \CommentTok{\# Remove extra newlines}
\NormalTok{    text }\OperatorTok{=}\NormalTok{ re.sub(}\VerbatimStringTok{r\textquotesingle{}\textbackslash{}n+\textquotesingle{}}\NormalTok{, }\StringTok{\textquotesingle{}}\CharTok{\textbackslash{}n}\StringTok{\textquotesingle{}}\NormalTok{, text)}

    \CommentTok{\# Remove page numbers associated }
    \CommentTok{\# with form feed characters (e.g., \textquotesingle{}\textbackslash{}x0c2\textquotesingle{})}
\NormalTok{    text }\OperatorTok{=}\NormalTok{ re.sub(}\VerbatimStringTok{r\textquotesingle{}\textbackslash{}x0c\textbackslash{}d+\textquotesingle{}}\NormalTok{, }\StringTok{\textquotesingle{}\textquotesingle{}}\NormalTok{, text)}

    \CommentTok{\# Fix hyphenated words across line breaks}
\NormalTok{    text }\OperatorTok{=}\NormalTok{ re.sub(}\VerbatimStringTok{r\textquotesingle{}{-}\textbackslash{}n\textquotesingle{}}\NormalTok{, }\StringTok{\textquotesingle{}\textquotesingle{}}\NormalTok{, text)}

    \CommentTok{\# Find "B.E." and remove punctuation}
\NormalTok{    text }\OperatorTok{=}\NormalTok{ re.sub(}\VerbatimStringTok{r\textquotesingle{}B\textbackslash{}.E\textbackslash{}.\textquotesingle{}}\NormalTok{, }\StringTok{\textquotesingle{}BE\textquotesingle{}}\NormalTok{, text)}

    \CommentTok{\# Remove extra spaces}
\NormalTok{    text }\OperatorTok{=}\NormalTok{ re.sub(}\VerbatimStringTok{r\textquotesingle{}\textbackslash{}s+\textquotesingle{}}\NormalTok{, }\StringTok{\textquotesingle{} \textquotesingle{}}\NormalTok{, text).strip()}

    \ControlFlowTok{return}\NormalTok{ text}

\CommentTok{\# Step 3: process all PDFs in the folder}
\KeywordTok{def}\NormalTok{ process\_pdfs\_in\_folder(folder\_path):}
    \CommentTok{"""}
\CommentTok{    Processes all PDF files in a folder to extract and clean text.}
\CommentTok{    Returns a pandas DataFrame with the document ID and cleaned text.}
\CommentTok{    """}
    \CommentTok{\# Initialize an empty list to hold extracted data}
\NormalTok{    data }\OperatorTok{=}\NormalTok{ [] }

    \ControlFlowTok{for}\NormalTok{ file\_name }\KeywordTok{in}\NormalTok{ os.listdir(folder\_path):}
        \ControlFlowTok{if}\NormalTok{ file\_name.lower().endswith(}\StringTok{".pdf"}\NormalTok{):}
\NormalTok{            pdf\_path }\OperatorTok{=}\NormalTok{ os.path.join(folder\_path, file\_name)}
            \CommentTok{\# uncomment if needed}
            \CommentTok{\#print(f"Processing: \{pdf\_path\}")}

            \CommentTok{\# Extract and clean text}
\NormalTok{            raw\_text }\OperatorTok{=}\NormalTok{ extract\_text\_from\_pdf(pdf\_path)}
            \ControlFlowTok{if}\NormalTok{ raw\_text:}
\NormalTok{                cleaned\_text }\OperatorTok{=}\NormalTok{ clean\_extracted\_text(raw\_text)}
\NormalTok{                data.append(\{}\StringTok{"text"}\NormalTok{: cleaned\_text\}) }

    \CommentTok{\# Step 4: create a DataFrame from the extracted data}
\NormalTok{    df }\OperatorTok{=}\NormalTok{ pd.DataFrame(data)}

    \CommentTok{\# Add an \textquotesingle{}id\textquotesingle{} column based on the index, starting from 1}
\NormalTok{    df[}\StringTok{\textquotesingle{}id\textquotesingle{}}\NormalTok{] }\OperatorTok{=}\NormalTok{ df.index }\OperatorTok{+} \DecValTok{1}

    \ControlFlowTok{return}\NormalTok{ df}

\CommentTok{\# Process PDFs and save to a DataFrame}
\NormalTok{df }\OperatorTok{=}\NormalTok{ process\_pdfs\_in\_folder(folder\_path)}
\end{Highlighting}
\end{Shaded}

\subsection{A1: Preprocessing Text for
Labeling}\label{a1-preprocessing-text-for-labeling}

The raw text is preprocessed to facilitate labeling by segmenting
documents into sentences using the NLTK library's
\texttt{sent\_tokenize} method. Each sentence is associated with its
document ID and assigned a placeholder for sentiment labels. To reduce
potential ordering bias, the sentences are randomized. The final
structured data is saved as a CSV file for annotation purposes.

\begin{Shaded}
\begin{Highlighting}[]
\CommentTok{\# Import function for splitting text into sentences }
\ImportTok{from}\NormalTok{ nltk.tokenize }\ImportTok{import}\NormalTok{ sent\_tokenize}

\CommentTok{\# Ensure nltk resources are downloaded}
\NormalTok{nltk.download(}\StringTok{\textquotesingle{}punkt\textquotesingle{}}\NormalTok{)}

\CommentTok{\# Step 1: segment each document into sentences}
\NormalTok{segmented\_data }\OperatorTok{=}\NormalTok{ []}
\ControlFlowTok{for}\NormalTok{ doc\_id, text }\KeywordTok{in} \BuiltInTok{zip}\NormalTok{(df[}\StringTok{\textquotesingle{}id\textquotesingle{}}\NormalTok{], df[}\StringTok{\textquotesingle{}document\textquotesingle{}}\NormalTok{]):}
\NormalTok{    sentences }\OperatorTok{=}\NormalTok{ sent\_tokenize(text)}
    \ControlFlowTok{for}\NormalTok{ sentence }\KeywordTok{in}\NormalTok{ sentences:}
\NormalTok{        segmented\_data.append(\{}
            \StringTok{"document\_id"}\NormalTok{: doc\_id, }\CommentTok{\#for reference if needed}
            \StringTok{"text"}\NormalTok{: sentence,}
            \StringTok{"sentiment"}\NormalTok{: }\VariableTok{None}  \CommentTok{\# Placeholder for annotation}
\NormalTok{        \})}

\CommentTok{\# Step 2: create a new DataFrame with segmented sentences}
\NormalTok{segmented\_df }\OperatorTok{=}\NormalTok{ pd.DataFrame(segmented\_data)}

\CommentTok{\# Step 3: shuffle the DataFrame to randomize the sentences}
\NormalTok{segmented\_df }\OperatorTok{=}\NormalTok{ segmented\_df.sample(frac}\OperatorTok{=}\DecValTok{1}\NormalTok{).reset\_index(drop}\OperatorTok{=}\VariableTok{True}\NormalTok{)}

\CommentTok{\# Step 4: save for annotation}
\NormalTok{segmented\_df.to\_csv(}\StringTok{\textquotesingle{}sentences\_for\_annotation.csv\textquotesingle{}}\NormalTok{, index}\OperatorTok{=}\VariableTok{False}\NormalTok{)}
\end{Highlighting}
\end{Shaded}

\section{Appendix B: Sentiment Annotation
Guidelines}\label{appendix-b-sentiment-annotation-guidelines}

This appendix outlines the criteria and process used for annotating
sentiment in the dataset.

\subsection{Sentiment Labels}\label{sentiment-labels}

\begin{enumerate}
\def\labelenumi{\arabic{enumi}.}
\tightlist
\item
  \textbf{Positive (1)}:

  \begin{itemize}
  \tightlist
  \item
    Sentences with an optimistic or favorable tone.
  \item
    Examples:

    \begin{itemize}
    \tightlist
    \item
      ``The Committee viewed that government measures should be
      expedited to support the economic recovery.''
    \item
      ``Growth is expected to pick up in 2024 due to steady tourism
      recovery and improved exports.''
    \end{itemize}
  \end{itemize}
\item
  \textbf{Negative (-1)}:

  \begin{itemize}
  \tightlist
  \item
    Sentences with a pessimistic or unfavorable tone.
  \item
    Examples:

    \begin{itemize}
    \tightlist
    \item
      ``Financial fragilities remain for some SMEs and households
      exposed to rising costs.''
    \item
      ``Core inflation remains elevated, posing risks to economic
      stability.''
    \end{itemize}
  \end{itemize}
\item
  \textbf{Neutral (0)}:

  \begin{itemize}
  \tightlist
  \item
    Fact-based or descriptive sentences without evident sentiment.
  \item
    Examples:

    \begin{itemize}
    \tightlist
    \item
      ``The document was published on 12th October.''
    \item
      ``The Committee projected economic growth of 3.6\% in 2023 and
      3.8\% in 2024.''
    \end{itemize}
  \end{itemize}
\end{enumerate}

\subsection{Annotation Process}\label{annotation-process}

The sentiment annotation involved a combination of automated and manual
steps:

\begin{itemize}
\tightlist
\item
  Initial annotations were generated using Generative AI (ChatGPT),
  prompted to classify sentences as positive, negative or neutral based
  on their tone and context.
\item
  The author, as a domain expert, reviewed and corrected AI-generated
  annotations to ensure accuracy, particularly aligning labels with the
  economic policy context.
\end{itemize}

\subsubsection{Prompt}\label{prompt}

The following prompt was used to generate sentiment annotations for the
sentences:

\begin{quote}
``Classify the sentiment of the following sentence into one of the three
categories: Positive, Negative or Neutral. Provide a brief explanation
for your classification.''
\end{quote}

\subsection{Quality Assurance}\label{quality-assurance}

To maintain high-quality annotations and minimize bias:

\begin{itemize}
\tightlist
\item
  Sentences were reviewed for tone consistency, factoring in contextual
  nuances.
\item
  Mixed sentiments were marked neutral, while ambiguous cases were
  carefully evaluated or defaulted to neutral as well.
\end{itemize}

\end{document}